\documentclass{article}

\usepackage{iclr2027_conference,times}

\usepackage{amsmath,amssymb,amsthm,mathtools}
\usepackage{booktabs}
\usepackage{longtable}
\usepackage{tabularx}
\usepackage{array}
\usepackage{graphicx}
\usepackage{wrapfig}
\usepackage{microtype}
\usepackage[hidelinks]{hyperref}
\usepackage{url}
\usepackage{xcolor}
\usepackage{tikz}
\usetikzlibrary{arrows.meta,positioning,fit,calc,decorations.pathreplacing}
\usepackage[capitalise,noabbrev]{cleveref}

\newcommand{\R}{\mathbb{R}}
\newcommand{\SA}{\operatorname{SA}}
\newcommand{\MHA}{\operatorname{MHA}}

\newcommand{\softmax}{\operatorname{softmax}}

\DeclareMathOperator*{\Emean}{\mathbb{E}}

\title{FTFM: Factorized Tabular Foundation Models}

\author{Josif Grabocka\\
University of Technology Nuremberg \\
\url{josif.grabocka@utn.de}}

\iclrfinalcopy

\begin{document}

\maketitle
\lhead{}
\renewcommand{\headrulewidth}{0pt}

\begin{abstract}
Tabular foundation models (TFMs) are becoming a powerful alternative to conventional tabular learning, adapting to new datasets in context without task-specific training. Current architectures, however, face a trade-off: models that preserve cell-level representations are expensive because they repeatedly attend across rows and columns, while faster models collapse each row into a fixed embedding and lose this fine-grained state.
We introduce Factorized Tabular Foundation Models (FTFM), which preserve cell representations while factorizing their computation through one contextual latent per row and one per column. Cells no longer attend to one another directly; instead, each cell is updated through the interaction of its row and column latents, reducing repeated attention while retaining feature-specific information.
FTFM achieves competitive predictive performance in tabular classification on major benchmarks such as TabArena and BeyondArena.
\end{abstract}

\begin{center}
\url{https://github.com/machinelearningnuremberg/ftfm-official}
\end{center}

\section{Introduction}
\label{sec:intro}

A tabular foundation model is a single network that reads a labeled training
set together with the test rows and returns predictions in one forward pass,
without a gradient step on the target dataset. Its parameters are meta-learned
once, over datasets drawn from a prior, and are then held fixed for every table
it is applied to (\cref{sec:icl}). Since TabPFN \citep{hollmann2023tabpfn}
showed that such a model outperforms tuned gradient-boosted ensembles, models
of this type have occupied the leading positions of the public benchmarks
\citep{tabpfn3,qu2026tabiclv2,eo2026exaone}.

Two architectural designs are in use by the state-of-the-art models, and both of them are variations of the classical Transformer architecture for tabular data. Feature-space preserving models \citep{hollmann2025tabpfn,eo2026exaone}
maintain a grid of cell representations and update it
with attention that alternates between rows and columns. These models force the row
attention to be executed once per column and the column attention once per row.
Therefore, for a table with $N$ rows and $M$ features at width $D$, one layer costs $\Theta(MN^2D +NM^2D + NMD^2)$, which is prohibitive on large tables. Feature-space collapsing models
\citep{qu2025tabicl,qu2026tabiclv2,tabpfn3,tabfm2026} compress each row into a
fixed-dimensional embedding before the in-context stage. The compression
layers, however, retain both row and column cell-grid attentions as pre-ICL feature
transformers. The attention over rows is computed through a small
set of learned inducing points and so is linear in $N$, but this is an
approximation of the attention it replaces rather than a cheaper exact
computation. In contrast, the attention over the $M$ features of a row is not reduced at all
and still costs $\Theta(NM^2D)$. Such a collapse is also a projection of the
feature space onto a static representation shared across all datasets, after
which no cell-level state remains, so any information the in-context stage
requires must pass a pre-ICL bottleneck whose width was fixed at pretraining time.
This practice therefore trades speed against representational loss.

We fix tabular models with a novel design, which we call factorized
tabular models (FTFM). At every layer, the table is summarized by one latent vector per row and
one per column, and each cell representation is updated as a factorization of
the latents of its own row and its own column through a learned interaction
(\cref{sec:factorized}). The construction generalizes classical matrix
factorization \citep{koren2009mf} and replaces cell-to-cell attention with row-to-row and column-to-column interaction. The new paradigm drops the per-layer cost to $\Theta(N^2D + M^2D + NMD^2)$, comparable to the feature-collapsing architectures. Therefore, our architecture offers (i) the expressivity of feature-preserving models, and (ii) the reduced inference runtime of the feature-collapsing architectures.

Furthermore, our modeling comes at no sacrifice in empirical predictive performance. We provide extensive experiments in terms of classification performance on the two most popular benchmarks for table prediction, namely TabArena~\citep{erickson2025tabarena} (38 datasets) and BeyondArena~\citep{purucker2026beyondarena} (98 datasets). FTFM achieves competitive performance on both benchmarks. Our model is a novel architecture for tabular data that is able to match state-of-the-art results.

We contribute (i) FTFM, a novel family of architectures based on deep table factorizations; (ii) a mechanism that preserves the feature space at the inference complexity of architectures that collapse features; and (iii) strong experiments on TabArena and BeyondArena rivaling the state-of-the-art foundation models.

\section{Related work}
\label{sec:related}

\paragraph{Tabular foundation models.}
Meta-learning one network across many tables predates the current wave of
models. The first tabular foundation model is Dataset2Vec
\citep{jomaa2021dataset2vec} (arXiv 2019; published 2021), a Deep-Set
architecture meta-learned over tables to produce latent representations of
whole datasets. In today's vocabulary, it is a self-supervised world model of
tables and not a supervised predictor, so it was never compared against a
classifier. \citet{iwata2020heterogeneous} introduced the first tabular
foundation model for supervised prediction, again built on Deep Sets. It did
not displace gradient-boosted trees
\citep{chen2016xgboost,prokhorenkova2018catboost}, which at the time still beat
deep models on mid-sized benchmarks \citep{grinsztajn2022trees}. TabPFN
\citep{hollmann2023tabpfn} was the first meta-learned supervised model to
outperform tuned tree ensembles on small classification benchmarks, using a
Transformer meta-learned over millions of synthetic datasets drawn from a prior
\citep{muller2022pfn}. Its successors held the state of the art for several
years, most recently TabPFN-3 \citep{tabpfn3}, which has since been rivaled
by TabFM \citep{tabfm2026} and EXAONE Tabular \citep{eo2026exaone}. Among fully
open models, where the architecture, the meta-learning code and the synthetic
pretraining data are all released, TabICL \citep{qu2025tabicl} and its
successor TabICLv2 \citep{qu2026tabiclv2} are the strongest.

The strongest models of this wave differ mainly in their treatment of the feature axis
and fall into two families. \emph{Feature-space preserving} models keep an $N \times
M$ grid of cell representations at every layer and update it by attention that
alternates between rows and columns, that is, axial attention
\citep{ho2019axial} over the table. The pattern originates with
\citet{lorch2022avici}, who introduced alternating attention over the
observation and variable axes of a data matrix for amortized causal structure
learning, and was popularized for prediction by TabPFN-2
\citep{hollmann2025tabpfn}, followed by TabPFN-2.5
\citep{grinsztajn2025tabpfn25}, EXAONE Tabular \citep{eo2026exaone} and LimiX
\citep{limix2025}. \emph{Feature-space collapsing} models instead project each row
onto a fixed number of shared feature dimensions and run in-context learning
directly across rows. TabICL established the two-stage form of a feature
representation stage followed by an ICL stage, and most current state-of-the-art models adopt it, among them TabPFN-3 and TabFM \citep{tabfm2026}. On the other hand, TabDPT \citep{ma2025tabdpt} is a single-stage variant that embeds whole rows as tokens and adds retrieval. There also exist LLM-based models, which serialize tables into text or build on pretrained language backbones~\citep{hegselmann2023tabllm,yan2024tpberta,gardner2024tabula}. They exploit column-name semantics that the other two families discard, but have so far not matched native tabular architectures on standard benchmarks~\citep{ma2025tabdpt}.

\paragraph{Matrix factorization.}
Latent-factor models of tables are older than tabular deep learning itself. In
collaborative filtering, a rating matrix is approximated by inner products of
row and column factors plus bias terms \citep{koren2009mf}. Under a low-rank
assumption, the missing entries are recoverable exactly
\citep{candes2009completion}. Factorization machines \citep{rendle2010fm}
extend the construction to arbitrary feature vectors by assigning every feature
a latent vector and scoring pairwise interactions with inner products. In all
of these models, the factors are free parameters, estimated once per dataset. We
retain the functional form of a cell explained by a row latent, a column latent
and their interaction, but compute the latents by attention over the table
itself. However, FTFM does not train a factorization of each table, but constructs it implicitly through in-context learning in a single pass.

\paragraph{Table factorizations.}
Neural networks that factorize tables have been explored by prior work. For instance, TabularNet \citep{du2021tabularnet} pools row-level and
column-level states, next to a graph encoder over cell relations. Similarly, neural factorization machines \citep{he2017nfm} model products of fitted feature embeddings through an MLP, and DCN V2 \citep{wang2021dcnv2} learns explicit
low-rank feature crosses in a similar setting. However, these methods train the factorized matrices separately for each given dataset. In contrast, FTFM is a foundation model that introduces a novel architecture for computing the factorization of a table in a single forward pass, through in-context learning.

\paragraph{Efficient attention and scalable serving.}
The cost of attention over tables has been reduced in several directions. Inducing
points \citep{lee2019set} bottleneck an attention layer through a small learned
set, and TabICLv2, TabFM, and TabPFN-3 apply them in their column stage. LimiX-2M
\citep{limix2m2026} considers the opposite direction and reports a low-rank
collapse of the cell representations in trained tabular models.
Serving-oriented methods change the computed function to make in-context
inference affordable, by compressing the context \citep{zabergja2026taco}, by
replacing softmax attention with a linear variant \citep{zeng2025tabflex}, or
by dropping cell-level state in favor of long-context row attention
\citep{tabdptturbo2026}. Our factorization requires no such approximation on
the cell grid, and the support-side computation is reusable across queries
exactly. 

\section{Proposed method}
\label{sec:method}

Matrix factorization decomposes a matrix $X \in \R^{N \times M}$ with $N$ rows
and $M$ columns via the product of a row-latent matrix $U \in \R^{N \times D}$,
a column-latent matrix $V \in \R^{M \times D}$, row biases $U^0 \in \R^N$, and
column biases $V^0 \in \R^M$ \citep{koren2009mf}. Each cell is approximated as
$X_{ij} \approx \langle U_i, V_j \rangle + U^0_i + V^0_j$. This section
develops that principle into an architecture for tabular in-context learning.
Tables are represented by row latents and column latents throughout the
network, and every cell representation is a factorization of the two.

\subsection{Tabular in-context learning}
\label{sec:icl}

A supervised tabular task is a tuple
$\mathcal{D} = (X^{\mathrm{tr}}, y^{\mathrm{tr}}, X^{\mathrm{te}},
y^{\mathrm{te}})$ with training inputs $X^{\mathrm{tr}} \in \R^{N^{\mathrm{tr}}
\times M}$, labels $y^{\mathrm{tr}} \in \mathcal{Y}^{N^{\mathrm{tr}}}$, and
test inputs $X^{\mathrm{te}} \in \R^{N^{\mathrm{te}} \times M}$, where
$\mathcal{Y} = \{1, \dots, K\}$ for classification and $\mathcal{Y} = \R$ for
regression. A tabular foundation model is a single network $f_\theta$ that maps
the labeled training set and the test inputs directly to test predictions,
$\hat y^{\mathrm{te}} = f_\theta(X^{\mathrm{tr}}, y^{\mathrm{tr}},
X^{\mathrm{te}})$, with no parameter update on the target dataset. Its weights
are meta-learned over a prior $p(\mathcal{D})$ over datasets, by minimizing
$\Emean_{\mathcal{D} \sim p(\mathcal{D})}\bigl[\mathcal{L}(\hat y^{\mathrm{te}},
y^{\mathrm{te}})\bigr]$ over $\theta$. Here $\mathcal{L}$ compares the
predictions for all test rows against $y^{\mathrm{te}}$, using the mean cross-entropy
for classification and a mean-squared error loss for regression, and $p(\mathcal{D})$ is
in practice a structural-causal-model prior from which datasets are sampled on
the fly \citep{muller2022pfn,qu2025tabicl}.
All knowledge about how to learn is contained in $\theta$. 

\subsection{Factorized table representations}
\label{sec:factorized}

Internally, $f_\theta$ operates on latent representations of every cell. After stacking
the $N = N^{\mathrm{tr}} + N^{\mathrm{te}}$ rows of a task into one table
$X \in \R^{N \times M}$, the network lifts it to a tensor
$Z^{(0)} \in \R^{N \times M \times D}$ and transforms it further through layers
$Z^{(\ell)}$, so that the representation of cell $(i,j)$ can capture the
interactions of row $i$ with column $j$. The architecture families of
\cref{sec:related} differ in how $Z^{(\ell+1)}$ is computed from
$Z^{(\ell)}$. Preserving models let cells attend to cells along both axes, and
collapsing models delete the cell axis after an initial pre-ICL projection.

Our construction proposes a different principle. At every layer $\ell$, the table is summarized by a
row-latent matrix $U^{(\ell)} \in \R^{N \times D}$ and a column-latent matrix
$V^{(\ell)} \in \R^{M \times D}$, and each cell is updated by a
factorization of its row and column latents,
\begin{equation}
Z^{(\ell+1)}_{ij} \;=\; Z^{(\ell)}_{ij} \;+\;
g^{(\ell)}\!\bigl(U^{(\ell)}_i,\, V^{(\ell)}_j\bigr),
\qquad
g(u, v) \;=\; \operatorname{MLP}\!\bigl(\bigl[\,u \odot v;\; u;\; v\,\bigr]\bigr),
\label{eq:factorized}
\end{equation}
where $[\cdot\,;\cdot\,;\cdot]$ denotes concatenation. Classical matrix
factorization is
the degenerate case of \cref{eq:factorized}. Summing the Hadamard term
recovers $\langle U_i, V_j \rangle$, so a single layer with $g(u,v) = \langle
u, v\rangle$ and the bias terms reproduces the decomposition above. Three
aspects differ from the classical case. The reconstruction has $D$ dimensions
per cell instead of one scalar. The interaction is learned and not fixed to an
inner product. And the factors are not free parameters fitted to one dataset,
but are computed by the network itself as follows:

\begin{align}
U^{(\ell)}_i &= \SA_{\mathrm{row}}\!\left(Z^{(\ell)}_{i,:}\right),
&&\text{pool row $i$'s $M$ cells with a learned query,} \label{eq:rowpool}\\
V^{(\ell)}_j &= \SA_{\mathrm{col}}\!\left(Z^{(\ell)}_{1:N^{\mathrm{tr}},\,j}\right),
&&\text{pool column $j$'s $N^{\mathrm{tr}}$ cells with a learned query,}
\label{eq:colpool}\\
U^{(\ell)} &\leftarrow U^{(\ell)} + \MHA\!\left(U^{(\ell)};\,
U^{(\ell)}_{1:N^{\mathrm{tr}}}\right),
&&\text{rows attend to training rows (ICL),} \label{eq:rowattn}\\
V^{(\ell)} &\leftarrow V^{(\ell)} + \MHA\!\left(V^{(\ell)};\, V^{(\ell)}\right),
&&\text{columns attend to columns,} \label{eq:colattn}
\end{align}
where $\SA$ denotes a pooling attention that summarizes a set of vectors
through a single learned query, and $\MHA\!\left(A;\, B\right)$ denotes
multi-head attention with queries from $A$ and keys and values from $B$.
The label $y^{\mathrm{tr}}_i$ enters additively into the cells of training row
$i$, so the row latents of \cref{eq:rowpool} are label-aware and the attention
of \cref{eq:rowattn} carries the in-context learning. The latent of a test row therefore
retrieves information from the latents of the labeled rows. 

\textbf{In FTFM, no cell ever attends directly to another cell.} All row-row interaction is carried by $N$ row latents and all column-column interaction by $M$ column latents, and the two are combined only
inside $g$. The cell grid and the feature space are preserved (i.e., $Z^{\ell} \in \R^{N \times M \times D}$), but the computation is factorized.


The prediction of our model is a function of the row latents and the training labels only, inspired by \citet{arbel2025equitabpfn}. Splitting the final
latents $U^{(L)}$ into their training and test rows, $U^{\mathrm{tr}}$ and
$U^{\mathrm{te}}$,
\begin{equation}
\hat y^{\mathrm{te}} \;=\; \frac{1}{H} \sum_{h=1}^{H}
\softmax\!\left(
  \frac{\bigl(U^{\mathrm{te}} W_q^{(h)}\bigr)
        \bigl(U^{\mathrm{tr}} W_k^{(h)}\bigr)^{\!\top}}{\sqrt{d_h}}
\right) y^{\mathrm{tr}},
\label{eq:dec}
\end{equation}
where $W_q^{(h)}, W_k^{(h)} \in \R^{D \times d_h}$ and the softmax is taken
over training rows; the prediction is computed as a head-averaged, attention-weighted mean of the training
labels. The labels enter as values of
the attention, one-hot encoded for classification and as the target value for
regression, so \cref{eq:dec} is the same operation for both task types, and the
network $f_\theta$ of \cref{sec:icl} is the composition of the cell embedder,
the $L$ factorized blocks, and \cref{eq:dec}.


\subsection{Asymptotic advantage of the factorization}
\label{sec:asymptotics}

The tabular foundation model families differ in the interaction terms, that is, in the attention scores.
Feature-space preserving models let the $N$ cells of a column attend to one
another, once per column, $\Theta(MN^2D)$, and the $M$ cells of a row attend to
one another, once per row, $\Theta(NM^2D)$. In addition, a feed-forward layer on each cell adds a $\Theta(NMD^2)$ complexity. A forward pass therefore costs
$\Theta(MN^2D + NM^2D + NMD^2)$ \citep{hollmann2025tabpfn,eo2026exaone}.

\begin{table}[h]
\centering
\scriptsize
\setlength{\tabcolsep}{3pt}
\caption{Asymptotic cost of one in-context forward pass}
\label{tab:asymptotics}
\begin{tabular}{@{}llcc@{}}
\toprule
Family & Examples & Inference complexity & Feature preservation \\
\midrule
Preserving & TabPFN-2, EXAONE Tabular & $\Theta(MN^2D + NM^2D + NMD^2)$ & yes \\
Collapsing & TabICL v1/v2, TabPFN-3, TabFM & $\Theta(N^2D + NM^2D + NMD^2)$ & \textbf{no} \\
Factorized (ours) & & $\Theta(N^2D + M^2D + NMD^2)$ & yes \\
\bottomrule
\end{tabular}
\end{table}

Feature-space collapsing models retain both of these attentions as pre-ICL
feature transformers and add an ICL stage over the collapsed row embeddings,
$\Theta(N^2D)$. In the pre-ICL stage, the attention over the $N$ cells of a
column is routed through $I$ learned inducing points, so those cells never
attend to one another directly and that attention costs $\Theta(NMID)$, which
is $\Theta(NMD)$ for constant $I$. The attention over the $M$
features of a row is not reduced and remains $\Theta(NM^2D)$. The feed-forward layer at the cell level adds a complexity of $\Theta(NMD^2)$. One forward pass costs
$\Theta(N^2D + NM^2D + NMD^2)$~\citep{qu2025tabicl,qu2026tabiclv2}.

In FTFM, the row latents attend to the training rows, $N$ of them in total rather
than once per column, $\Theta(N^2D)$, and the column latents attend to the
column latents, $M$ in total rather than once per row, $\Theta(M^2D)$. We also apply cell feed-forward layers for a complexity of $\Theta(NMD^2)$. One forward pass costs $\Theta(N^2D + M^2D + NMD^2)$. FTFM has an asymptotic edge over other feature-preserving architectures, and is comparable to feature-collapsing models. Therefore, it maintains the feature space expressivity at the same asymptotic cost as feature-collapsing architectures.

\section{Experiments}
\label{sec:experiments}

\subsection{Experimental protocol}
\label{sec:protocol}

\paragraph{Model and pretraining.} Our FTFM model has $121.6$M
parameters and consists of $16$ factorized blocks of width $D=512$ with
$H=8$ heads, feature groups of $3$, robust per-column normalization, and the
equivariant vote head.  We meta-learn FTFM following a three-stage curriculum, each stage warm-started from the one
before it (\cref{tab:curriculum}). Stage~A adopts TabICLv2's stage-1 regime
\citep{qu2026tabiclv2}: their graph structural-causal-model prior, their
envelope of shapes and class counts, and their $500$k updates. Stages~B and~C draw each update's batch from that prior or, with equal probability, from self-supervised episodes over the corpus
of real tables used by TabDPT \citep{ma2025tabdpt}, while widening the tables and
taking the support from a fixed $1{,}024$ rows to a maximum of $32{,}768$.
\Cref{app:priors} describes both arms in more detail.

\begin{table}[h]
\centering
\small
\caption{The pretraining curriculum of FTFM}
\label{tab:curriculum}
\begin{tabular}{@{}llccrcr@{}}
\toprule
Stage & Prior & Rows & Features & Updates & Peak LR & GPU-h \\
\midrule
A & Synthetic only & $1{,}024$ & $1$--$100$ & $500$k & $8\times 10^{-4}$ & $1{,}600$ \\
B & Synthetic and real; $50\%-50\%$ & $256$--$16{,}384$ & $4$--$128$ & $200$k & $8\times 10^{-5}$ & $540$ \\
C & Synthetic and real; $50\%-50\%$ & $256$--$32{,}768$ & $4$--$128$ & $130$k & $1\times 10^{-5}$ & 288 \\
\bottomrule
\end{tabular}
\end{table}

At inference time, we ensemble FTFM by perturbing the sign of numerical features, randomly perturbing the categorical codes, and randomly applying power transformations to features. We do not shuffle the feature order since our method does not use RoPE and is fully permutation invariant. We create an ensemble of 8 base models, each fed with a perturbed variant of the input table.

\paragraph{Optimization.} We use the same training pipeline for all stages, apart from the prior hyperparameters described in~\cref{tab:curriculum}. As an optimizer, we deploy Muon \citep{liu2025muon} on every matrix parameter, with
momentum $0.9$, Nesterov acceleration, five Newton--Schulz steps, decoupled weight decay
$0.01$, gradients clipped at $10$ and non-finite ones skipped. The schedule is
a linear warmup over the first $1\%$ of updates and a cosine decay to $\frac{1}{8\times10^{-4}}$
of the peak in stage~A and $\frac{1}{8\times10^{-3}}$ in stages~B and~C. Batch size is not fixed:
each update packs tables until an activation-memory budget is met, capped at
$48$ tables in stage~A and $128$ afterwards, with table shapes synchronized across
data-parallel ranks. Stages~B and~C keep an exponential moving average of the
weights at $0.999$, and it is that average which is released and evaluated.

\paragraph{Evaluation and baselines.} We study \emph{classification} in this paper as a first experimental framework for verifying the feasibility of factorized table models. We
evaluate on two benchmarks, both of which pin their datasets, splits, and
metrics: TabArena \citep{erickson2025tabarena}, whose classification suite is
$38$ datasets, and BeyondArena \citep{purucker2026beyondarena}, which adds
non-IID, temporal, grouped, and size-stressed slices over $98$ classification datasets and on
which foundation models have not been leading so far.

As baselines, we take the strongest representative of each architecture family,
and they span two orders of magnitude in size. TabFM
\citep{tabfm2026} is the largest at $1.64$B parameters, a collapsing zero-shot
model released as code and weights with no technical report, so our account of
its architecture comes from the released implementation. TabICLv2 \citep{qu2026tabiclv2}, the strongest fully open model and a two-stage
collapsing architecture, is $27.6$M, of which $26.3$M sit in a deep stack that
runs after the feature axis has been collapsed away. EXAONE Tabular
\citep{eo2026exaone} is smaller still at $20.81$M and is feature-space
preserving. TabPFN-3 \citep{tabpfn3} is the strongest of the TabPFN line, a
collapsing architecture having 53.2M parameters that scales to $10^6$ training rows and adds test-time
compute scaling. Tuned gradient-boosted trees \citep{chen2016xgboost,prokhorenkova2018catboost}
serve as the classical reference. 

\begin{figure}[t]
\centering
\includegraphics[width=\textwidth]{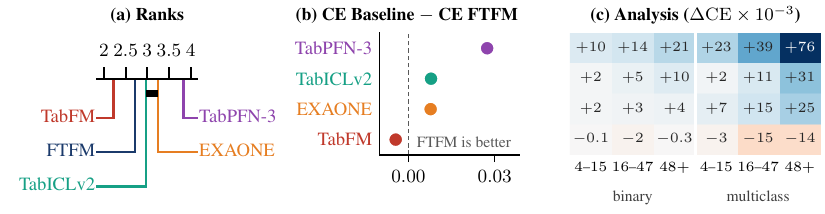}
\caption{Performance on $4$K held-out tables
from stage 1 of the TabICLv2 prior, with every model scored on the same tables. (a) Average ranks; a bar indicates a non-significant difference following a Wilcoxon signed-rank test; (b) mean paired $\Delta$CE (cross-entropy) against FTFM; (c) analysis by number of features and number of classes.}
\label{fig:rq1_paired}
\end{figure}

\subsection{Research Questions (RQs) and Associated Experiments}
\label{sec:rqs}

\subsubsection{RQ1: Are factorized tabular models expressive enough?}
\label{sec:rq1}

Since FTFM is a novel paradigm for tabular foundation models that does not apply any direct cell-to-cell attention, one might suspect that it introduces a restrictive expressivity bias. To investigate this hypothesis, we test the ability of FTFM to fit a prior in a controlled experiment. We meta-learn FTFM following the Stage-A curriculum specified in Table~\ref{tab:curriculum}, which is the exact prior and number of update steps that TabICLv2 was pretrained on. If FTFM outperforms TabICLv2 on its own prior and meta-learning regime of Stage-A, this indicates FTFM does not have obviously restrictive model expressivity.

The empirical evidence presented in \Cref{fig:rq1_paired} demonstrates that FTFM fits the same prior better than the TabICLv2 model, at the same budget.  In addition, it shows that on the prior FTFM is learned on, our model outperforms EXAONE Tabular and TabPFN-3 by a statistically significant margin, and is second only to the $10\times$-larger TabFM model. We deduce that our modeling of tables does not have an expressivity bias.

\subsubsection{RQ2: Do the meta-learning stages lead to progressive gains?}
\label{sec:rq2}

\begin{wrapfigure}{l}{0.50\textwidth}
\vspace{-\intextsep}
\centering
\includegraphics[width=0.48\textwidth]{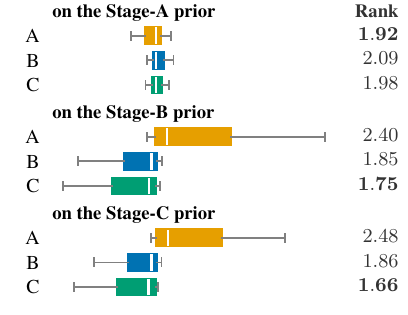}
\caption{Cross in-distribution evaluation of the pretraining stages. }
\label{fig:rq2_priors}
\vspace{-0.6cm}
\end{wrapfigure}

As a second question, we validate whether the choice and sequence of the three pretraining stages of \cref{tab:curriculum} lead to any meaningful benefit.  To test this question, we first evaluate each stage's checkpoint on the TabArena
Lite version~\citep{erickson2025tabarena}.

\begin{wraptable}{r}{0.48\textwidth}
\vspace{-0.7cm}
\centering
\small
\caption{The performance of each pretraining stage evaluated out-of-distribution on TabArena Lite. Significance is a
two-sided Wilcoxon signed-rank test.}
\label{tab:rq2}
\begin{tabular}{@{}lccc@{}}
\toprule
Stage & Mean error & C wins on & $p (C \text{ vs.})$ \\
\midrule
A & $0.1692$ & $30/38$  & $<0.001$ \\
B & $0.1620$ & $24/38$ & $0.030$ \\
C & $\mathbf{0.1614}$ & --- & --- \\
\bottomrule
\end{tabular}
\vspace{-0.3cm}
\end{wraptable}

All checkpoints are under one serving policy so that only the pretraining differs. None of the TabArena or BeyondArena datasets occur in any pretraining datasets. The results on TabArena Lite are shown in Table~\ref{tab:rq2}. We notice that the larger tables and the real datasets of stage B give a tangible lift in performance. Ultimately, the checkpoint of stage C, which is trained for a longer budget and on bigger tables, results in better out-of-distribution performance.

Furthermore, we check the in-distribution performance of the per-stage checkpoints, evaluating all stages on the specific prior of each stage. We score
all three checkpoints on $2$K held-out tables from each of the three
stage priors in turn. The results are presented in \Cref{fig:rq2_priors}. We observe that meta-learning on the bigger tables of stages B and C non-significantly deteriorates the performance on the small table prior of Stage A. However, the later stages intuitively give a lift on the larger tables they are meta-learned from.

\subsubsection{RQ3: Can FTFM reach the state of the art?}
\label{sec:rq3}

We highlight that the actual empirical research in the field of tabular foundation models compares models using benchmark performances. However, the considered methods are pretrained on different priors and with different training budgets, use different feature preprocessing of the datasets and different feature representation layers in the architectures, have different capacities, and are even ensembled differently at inference time. This practice makes it hard to assess whether the benefit of a released model comes from the architecture, especially if the pretraining code and data are not released for ablating the components. We also follow such performance-oriented practice in this paper, since ablating the benefit of our architecture would demand retraining the baseline models on our prior at the same budget and with the same feature preprocessing and ensembling, which is infeasible within our available computational budget.

Nevertheless, a positive standard in the field of tabular foundation models is the streamlined benchmarks, the most popular of which are TabArena and BeyondArena. In this section, we present the empirical performance of our method on these two benchmarks.

\begin{figure}[!h]
\centering
\includegraphics[width=0.96\textwidth]{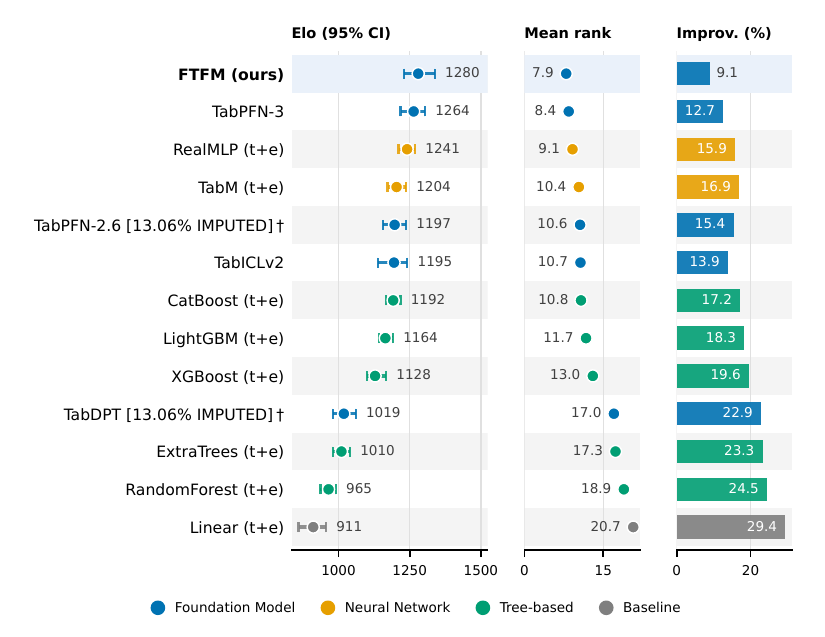}
\caption{Performance of FTFM on the 98 classification datasets of the BeyondArena benchmark. FTFM leads TabPFN-3 and TabICLv2 on Elo and on mean rank, and is lowest on improvability, which is the margin by which a method is beaten. The ranks are computed against 29 methods covered in the benchmark paper.}
\label{fig:rq3}
\end{figure}

\subsubsection*{Results on the BeyondArena benchmark}
In BeyondArena, our method achieves state-of-the-art performance as indicated in Figure~\ref{fig:rq3}. FTFM is first of $29$ methods: ahead of TabPFN-3, TabICLv2 and TabPFN-2.6 on Elo and on mean rank, and lowest of every entrant on improvability. Every row in that figure, ours and the
published ones alike, is scored by the benchmark's own context on exactly the
datasets and splits we ran. Furthermore, we detail the performance across the types of datasets in Table~\ref{tab:beyondarena_full}.

\begin{table}[h]
\centering
\small
\caption{BeyondArena Elo by slice on the classification datasets of each slice
that our run covers, with the place out of $29$ methods in parentheses and $n$
the number of datasets. Every column is recomputed on those same tasks with the
benchmark's own evaluator, so a row compares five methods on one set of
datasets.}
\label{tab:beyondarena_full}
\begin{tabular}{@{}lrccccc@{}}
\toprule
Slice & $n$ & TabPFN-3 & TabICLv2 & RealMLP & CatBoost & FTFM \\
      &      & (default) & (default) & (t+e) & (t+e) &  \\
\midrule
tiny & 39 & $1366$ (2) & $1333$ (3) & $1287$ (5) & $1233$ (9) & $\bf 1426$ (1) \\
small & 26 & $1408$ (2) & $1323$ (4) & $1270$ (5) & $1196$ (7) & $\bf 1510$ (1) \\
medium & 20 & $1201$ (2) & $1121$ (10) & $\bf 1201$ (1) & $1180$ (3) & $1168$ (4) \\
large & 13 & $912$ (16) & $668$ (21) & $\bf 1285$ (1) & $1237$ (5) & $772$ (17) \\
low-dim & 66 & $1334$ (2) & $1312$ (3) & $1261$ (5) & $1222$ (6) & $\bf 1390$ (1) \\
high-dim & 32 & $1139$ (4) & $983$ (17) & $\bf 1212$ (1) & $1140$ (3) & $1095$ (11) \\
high-cardinality & 21 & $1068$ (7) & $937$ (14) & $1146$ (4) & $\bf 1179$ (1) & $930$ (15) \\
random & 77 & $1341$ (2) & $1273$ (3) & $1251$ (5) & $1202$ (7) & $\bf 1361$ (1) \\
grouped & 11 & $1055$ (22) & $1097$ (16) & $\bf 1279$ (2) & $1215$ (5) & $1206$ (7) \\
temporal & 10 & $1051$ (9) & $755$ (19) & $\bf 1184$ (1) & $1145$ (4) & $866$ (17) \\
text & 9 & $1200$ (3) & $1064$ (13) & $\bf 1286$ (1) & $1217$ (2) & $998$ (18) \\
\midrule
all & 98 & $1264$ (2) & $1195$ (6) & $1241$ (3) & $1192$ (7) & $\bf 1280$ (1) \\
\bottomrule
\end{tabular}
\end{table}

The results of Table~\ref{tab:beyondarena_full} indicate that FTFM is the best method on tiny, small, and low-dim datasets, as well as those having random effects, while overall leading with an Elo of 1280. This matches the expectations since our prior does not have datasets with high-cardinality, grouped features, or temporal aspects. Overall, we deduce that FTFM offers a novel architecture that can be trained to the performance standard of the state-of-the-art closed and open tabular foundation models, and clearly outperforms traditional classifiers such as RealMLP or CatBoost.

\subsubsection*{Results on the TabArena benchmark}

\begin{wrapfigure}{r}{0.51\textwidth}
\vspace{-0.5cm}
\centering
\includegraphics[trim={0.5cm 0 0 0}, clip=true,width=0.5\textwidth]{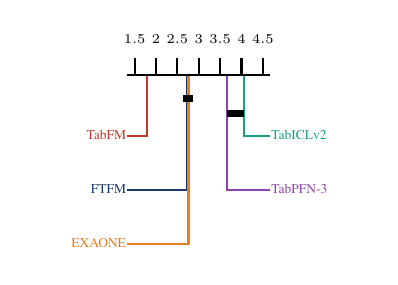}
\vspace{-1.0cm}
\caption{Critical-difference diagram over the full
TabArena protocol, $38$ datasets. Average rank, lower is better; a bar joins
methods whose performance differences are not statistically significant. In contrast to Figure~\ref{fig:tabarena_leaderboard}, here the ranks are computed only among the five selected methods.}
\vspace{-0.2cm}
\label{fig:tabarenafull_cd}
\end{wrapfigure}

We also evaluate FTFM on the TabArena benchmark, a living benchmark consisting of 38 classification datasets. The rank comparisons of the leading methods are presented in Figure~\ref{fig:tabarenafull_cd}. In the appendix, we also provide a leaderboard plot in Figure~\ref{fig:tabarena_leaderboard} and the detailed results in Table~\ref{tab:tabarena_full}. Overall, FTFM rivals the top foundation models and ranks among the top 3 methods out of 72 baselines evaluated in TabArena.

\begin{figure}[t]
\centering
\begin{minipage}[c]{0.62\textwidth}
\centering
\includegraphics[width=\textwidth]{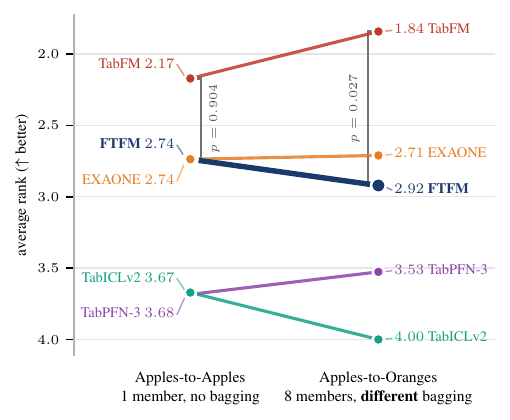}
\end{minipage}\hfill
\begin{minipage}[c]{0.35\textwidth}
\caption{Average rank over the $38$ TabArena Lite
datasets, apples-to-apples (left) and apples-to-oranges (right); each line is one
method moving between the regimes. On the right, every method is bagged over $8$
members, but each brings its own \textbf{different} ensembling pipeline. The
bracket carries the two-sided Wilcoxon signed-rank test of FTFM against TabFM,
Holm-corrected over the ten pairs. The left
column shows our runs and the right takes the four baselines from the benchmark's published values on identical dataset splits.}
\label{fig:ensembling_impact}
\end{minipage}
\end{figure}

We further highlight that the empirical performance of tabular foundation models depends on the quality of the inference-time ensembling. We conducted an experiment on the TabArena Lite benchmark, where we first ran an apples-to-apples experiment where each model is run without ensembling by receiving a single z-normalized table. Then we compare in an apples-to-oranges manner the ensembled variants of the methods (each with a different ensembling strategy) and present the results in Figure~\ref{fig:ensembling_impact}. FTFM is not worse than TabFM with a statistically significant margin on the apples-to-apples experiment, indicating that that part of the reported performance of models comes from their efficient ensembling.

\begin{wrapfigure}{l}{0.50\textwidth}
\centering
\includegraphics[width=0.48\textwidth]{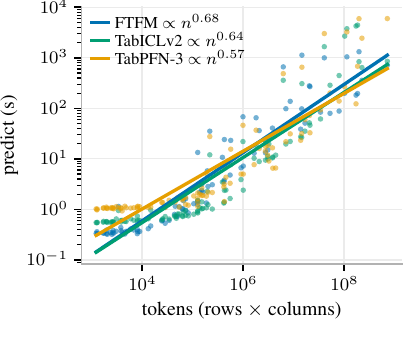}
\caption{Prediction time against table size on the BeyondArena classification
datasets.}
\label{fig:predict_scaling}
\vspace{-0.5cm}
\end{wrapfigure}

\vspace{0.3cm}
\subsection{Inference costs}

As a last remark, in Section~\ref{sec:asymptotics} we hypothesized that FTFM has the asymptotic inference costs of feature-collapsing baselines, even though it preserves the feature space. To validate this claim, we present in Figure~\ref{fig:predict_scaling} the prediction times of our method compared to TabICLv2 and TabPFN-3 on the 98 datasets of the BeyondArena benchmark. We plot the number of tokens in each table versus the prediction time in seconds. As observed, our method shows a linear log-log relationship between inference cost and data size, comparable to feature-collapsing methods. We present the full fit and prediction results for all the methods in Table~\ref{tab:beyondarena_cost} in the appendix.

These findings show that our novel factorized table modeling can achieve state-of-the-art performance while also offering practical benefits to practitioners by matching the speed of the fastest feature-collapsing tabular foundation models.

\section{Limitations}
\label{sec:limitations}

To validate the novel direction of factorized table representations, we focused on the classification task, leaving regression as future work. Furthermore, our architecture can benefit from the recent Fourier feature representation introduced in \citet{tabfm2026}.

\section{Conclusion}

We introduced FTFM, a tabular foundation model in which a cell representation
is a factorization of one contextual row latent against one column latent
rather than the product of cell-to-cell attention. Our method achieves strong performance on the BeyondArena and TabArena benchmarks, outperforming state-of-the-art methods in the field, such as TabPFN-3 and TabICLv2. We showcased that FTFM, a novel architecture based on table factorization,  achieves competitive performance in modern tabular representation learning and can serve as the future architectural blueprint for efficiently modeling tables.

\newpage

\section*{Reproducibility statement}

The architecture is specified operation by operation in
\cref{sec:factorized}. The overall design choices and hyperparameters are detailed in the paper. Our evaluation uses the published fixed protocols of TabArena and BeyondArena, which pin datasets, splits, and metrics. The synthetic pretraining prior follows the structural-causal-model family of
\citet{qu2025tabicl,qu2026tabiclv2} and the datasets of \citet{ma2025tabdpt}. Inference code and a link to the model weights accompany the submission.

\section*{AI Use Statement}

We used generative AI to prototype the source code for the experiments, followed by manual inspection and testing of the generated code. We also used AI to check for grammar issues and rephrase the manuscript text. Furthermore, generative AI models were used to generate and polish the visual aspects of the figures and tables. We did not use AI to generate the core idea, suggest experiments, baselines, datasets, or write the proposed method. We reviewed all AI-assisted work.

\bibliographystyle{iclr2027_conference}
\bibliography{ftfm}

\appendix
\newpage

\section{Full-protocol TabArena leaderboard}
\label{app:tabarena_leaderboard}

\begin{figure}[!htb]
\centering
\includegraphics[width=0.96\textwidth]{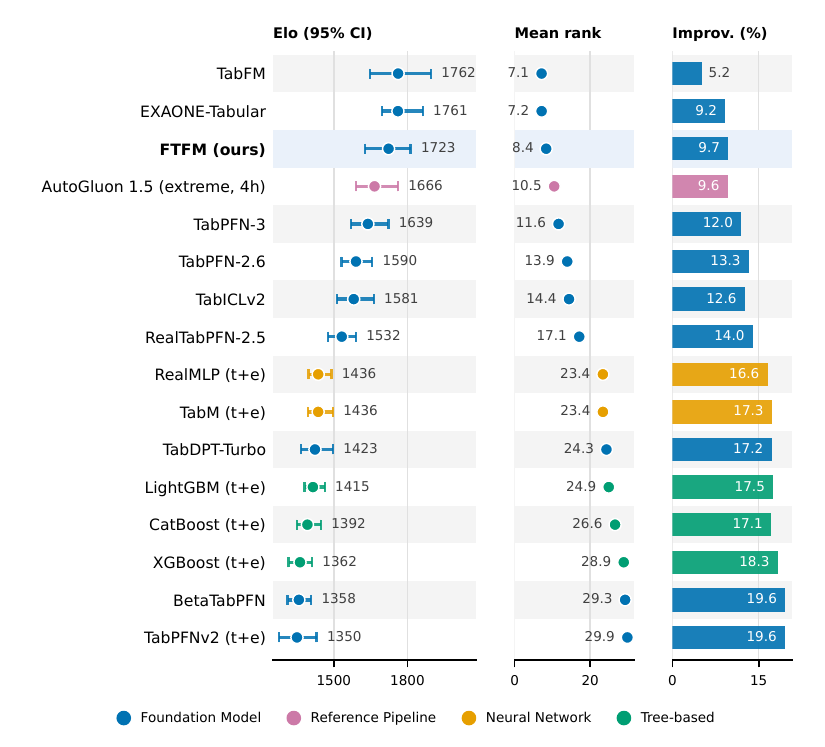}
\caption{TabArena classification on $38$ datasets and $594$ splits under the
full protocol. Results are recomputed with TabArena's evaluator from published
per-split results. Ranks are computed against the benchmark's $72$ published
methods, with selected methods shown here.}
\label{fig:tabarena_leaderboard}
\end{figure}

\Cref{fig:tabarena_leaderboard} complements the BeyondArena results in
\cref{fig:rq3}, using the same plot format.

\section{Per-dataset TabArena results}
\label{app:tabarena_full}

\Cref{tab:tabarena_full} reports the per-dataset results behind
\cref{fig:tabarena_leaderboard}, using published baseline results.

\begin{table}[h]
\centering
\scriptsize
\setlength{\tabcolsep}{5pt}
\caption{Error on the $38$ TabArena classification datasets under the full
protocol, averaged over splits. The metric is $1-\mathrm{AUC}$ for binary
tasks and log-loss for multiclass tasks. Lower is better, and the best result
in each row is bold.
Baselines use published per-split results.}
\label{tab:tabarena_full}
\begin{tabular}{@{}lccccc@{}}
\toprule
Dataset & TabICLv2 & TabPFN-3 & EXAONE & TabFM & FTFM \\
\midrule
APSFailure & $0.0058$ & $0.0061$ & $0.0061$ & $\mathbf{0.0054}$ & $0.0062$ \\
Amazon\_employee\_access & $0.1469$ & $0.1499$ & $\mathbf{0.1285}$ & $0.1420$ & $0.1398$ \\
Bank\_Customer\_Churn & $0.1281$ & $0.1248$ & $0.1253$ & $\mathbf{0.1233}$ & $0.1253$ \\
Bioresponse & $0.1382$ & $0.1228$ & $0.1185$ & $\mathbf{0.1136}$ & $0.1237$ \\
Diabetes130US & $0.3366$ & $0.3296$ & $\mathbf{0.3186}$ & $0.3299$ & $0.3327$ \\
E-CommereShippingData & $0.2549$ & $\mathbf{0.2529}$ & $0.2532$ & $0.2574$ & $0.2532$ \\
Fitness\_Club & $0.1792$ & $\mathbf{0.1785}$ & $0.1794$ & $0.1788$ & $0.1790$ \\
GiveMeSomeCredit & $0.1329$ & $0.1316$ & $0.1316$ & $\mathbf{0.1312}$ & $0.1325$ \\
HR\_Analytics\_Job\_Change\_of\_Data\_Scientists & $0.1938$ & $0.1939$ & $\mathbf{0.1930}$ & $0.1933$ & $0.1937$ \\
Is-this-a-good-customer & $0.2519$ & $0.2487$ & $\mathbf{0.2431}$ & $0.2464$ & $0.2461$ \\
MIC & $0.4454$ & $0.4449$ & $\mathbf{0.4248}$ & $0.4306$ & $0.4275$ \\
Marketing\_Campaign & $0.0752$ & $0.0838$ & $0.0791$ & $\mathbf{0.0736}$ & $0.0752$ \\
NATICUSdroid & $0.0122$ & $0.0125$ & $0.0123$ & $\mathbf{0.0114}$ & $0.0123$ \\
SDSS17 & $0.0714$ & $\mathbf{0.0641}$ & $0.0682$ & $0.0693$ & $0.0723$ \\
anneal & $0.0201$ & $0.0159$ & $\mathbf{0.0108}$ & $0.0132$ & $0.0143$ \\
bank-marketing & $0.2369$ & $0.2334$ & $0.2324$ & $\mathbf{0.2306}$ & $0.2332$ \\
blood-transfusion-service-center & $0.2446$ & $0.2460$ & $0.2504$ & $0.2441$ & $\mathbf{0.2434}$ \\
churn & $0.0648$ & $0.0662$ & $0.0659$ & $0.0666$ & $\mathbf{0.0578}$ \\
coil2000\_insurance\_policies & $0.2264$ & $0.2275$ & $0.2173$ & $\mathbf{0.2122}$ & $0.2234$ \\
credit-g & $0.2036$ & $0.2066$ & $0.1964$ & $\mathbf{0.1942}$ & $0.1955$ \\
credit\_card\_clients\_default & $0.2089$ & $0.2098$ & $0.2093$ & $\mathbf{0.2069}$ & $0.2077$ \\
customer\_satisfaction\_in\_airline & $0.0048$ & $0.0046$ & $0.0043$ & $\mathbf{0.0038}$ & $0.0050$ \\
diabetes & $0.1608$ & $0.1591$ & $0.1603$ & $0.1579$ & $\mathbf{0.1578}$ \\
hazelnut-spread-contaminant-detection & $0.0051$ & $0.0054$ & $0.0064$ & $\mathbf{0.0023}$ & $0.0048$ \\
heloc & $0.1985$ & $0.1977$ & $0.1975$ & $0.1974$ & $\mathbf{0.1973}$ \\
hiva\_agnostic & $0.1986$ & $0.1799$ & $\mathbf{0.1746}$ & $0.1891$ & $0.1852$ \\
in\_vehicle\_coupon\_recommendation & $0.1562$ & $0.1542$ & $0.1550$ & $0.1526$ & $\mathbf{0.1488}$ \\
jm1 & $0.2264$ & $0.2275$ & $0.2177$ & $\mathbf{0.2106}$ & $0.2205$ \\
kddcup09\_appetency & $0.1698$ & $0.1750$ & $\mathbf{0.1501}$ & $0.1549$ & $0.1664$ \\
maternal\_health\_risk & $0.3972$ & $0.4037$ & $0.4146$ & $\mathbf{0.3706}$ & $0.3857$ \\
online\_shoppers\_intention & $0.0625$ & $0.0614$ & $0.0611$ & $\mathbf{0.0601}$ & $0.0611$ \\
polish\_companies\_bankruptcy & $0.0156$ & $0.0150$ & $0.0095$ & $\mathbf{0.0047}$ & $0.0071$ \\
qsar-biodeg & $0.0591$ & $0.0610$ & $0.0616$ & $\mathbf{0.0580}$ & $0.0591$ \\
seismic-bumps & $0.2178$ & $0.2156$ & $0.2142$ & $\mathbf{0.2061}$ & $0.2123$ \\
splice & $0.0997$ & $0.1060$ & $0.0988$ & $0.0961$ & $\mathbf{0.0956}$ \\
students\_dropout\_and\_academic\_success & $0.5335$ & $0.5397$ & $0.5175$ & $\mathbf{0.5078}$ & $0.5255$ \\
taiwanese\_bankruptcy\_prediction & $0.0537$ & $0.0519$ & $0.0526$ & $\mathbf{0.0491}$ & $0.0506$ \\
website\_phishing & $0.2226$ & $0.2173$ & $0.2164$ & $\mathbf{0.2101}$ & $0.2227$ \\
\midrule
Mean & $0.1674$ & $0.1664$ & $0.1625$ & $0.1607$ & $0.1632$ \\
Mean rank & $4.05$ & $3.66$ & $2.76$ & $1.79$ & $2.74$ \\
\bottomrule
\end{tabular}
\end{table}

\section{Results across pretraining stages}
\label{app:prior_cross}

\Cref{tab:prior_cross} reports the values in \cref{fig:rq2_priors}. All
checkpoints are evaluated on the same $2{,}000$ held-out tables within each
prior. Compare checkpoints within rows because the priors differ in difficulty.

\begin{table}[h]
\centering
\small
\caption{Mean cross-entropy and average rank by checkpoint and evaluation
prior. The best value for each metric within each prior is bold. On the
Stage-B prior, different checkpoints minimize mean error and rank. The lower
panel gives two-sided Wilcoxon signed-rank $p$-values, Holm-corrected across
the three pairs within each prior.}
\label{tab:prior_cross}
\begin{tabular}{@{}lcccccc@{}}
\toprule
& \multicolumn{2}{c}{Stage-A} & \multicolumn{2}{c}{Stage-B} & \multicolumn{2}{c}{Stage-C} \\
\cmidrule(lr){2-3}\cmidrule(lr){4-5}\cmidrule(lr){6-7}
Scored on & CE & rank & CE & rank & CE & rank \\
\midrule
Stage-A & $\mathbf{0.2866}$ & $\mathbf{1.92}$ & $0.2883$ & $2.09$ & $0.2878$ & $1.98$ \\
Stage-B & $0.3566$ & $2.40$ & $\mathbf{0.3278}$ & $1.85$ & $0.3286$ & $\mathbf{1.75}$ \\
Stage-C & $0.4096$ & $2.48$ & $0.3900$ & $1.86$ & $\mathbf{0.3900}$ & $\mathbf{1.66}$ \\
\midrule
Scored on & \multicolumn{2}{c}{A vs B} & \multicolumn{2}{c}{A vs C} & \multicolumn{2}{c}{B vs C} \\
\midrule
Stage-A & \multicolumn{2}{c}{$6.9\times 10^{-8}$} & \multicolumn{2}{c}{$0.001$} & \multicolumn{2}{c}{$2.3\times 10^{-5}$} \\
Stage-B & \multicolumn{2}{c}{$1.0\times 10^{-102}$} & \multicolumn{2}{c}{$2.8\times 10^{-113}$} & \multicolumn{2}{c}{$0.033$} \\
Stage-C & \multicolumn{2}{c}{$1.1\times 10^{-126}$} & \multicolumn{2}{c}{$3.7\times 10^{-143}$} & \multicolumn{2}{c}{$8.4\times 10^{-14}$} \\
\bottomrule
\end{tabular}
\end{table}

\clearpage
\section{Per-dataset BeyondArena results}
\label{app:beyondarena_per_dataset}

\Cref{tab:beyondarena_error,tab:beyondarena_cost} report errors and runtimes
for all $98$ BeyondArena classification datasets. FTFM timings use one H100,
while baselines use the benchmark's hardware, so the times are not directly
comparable.

{\scriptsize
\setlength{\tabcolsep}{4pt}
\begin{longtable}{@{}lccccc@{}}
\caption{Error on the $98$ BeyondArena classification datasets, averaged over
splits. Lower is better, and the best result in each row is bold. Baselines use
published per-split results. $^{\dagger}$ marks FTFM's two reduced-support
results ($500{,}000$ rows, rung~$3$), which are not directly comparable to
full-support results (see \cref{app:serving_policy}).}
\label{tab:beyondarena_error} \\
\toprule
Dataset & FTFM & TabPFN-3 & TabICLv2 & RealMLP & CatBoost \\
\midrule
\endfirsthead
\toprule
Dataset & FTFM & TabPFN-3 & TabICLv2 & RealMLP & CatBoost \\
\midrule
\endhead
\midrule
\multicolumn{6}{r}{\emph{continued on the next page}} \\
\endfoot
\bottomrule
\endlastfoot
acquire\_valued\_shoppers\_challenge-f & $0.3518$ & $\mathbf{0.3468}$ & $0.3681$ & $0.3516$ & $0.3559$ \\
amazon\_employee\_access-4417c4c952bc & $0.1318$ & $0.1489$ & $0.1419$ & $0.1276$ & $\mathbf{0.1134}$ \\
amex\_non\_iid\_1m-e2e64ee8b42e & $0.0574$ & $0.0618$ & $0.0631$ & $\mathbf{0.0546}$ & $0.0559$ \\
anes\_voting\_2026-58a0e941922e & $0.1290$ & $\mathbf{0.1200}$ & $0.1305$ & $0.1283$ & $0.1272$ \\
aps\_failure-b63fc6330117 & $0.0067$ & $\mathbf{0.0060}$ & $0.0062$ & $0.0083$ & $0.0086$ \\
asp\_potassco\_classification-6cbac33e & $2.2139$ & $2.2067$ & $\mathbf{2.1993}$ & $2.2158$ & $2.2245$ \\
audiology\_diagnosis-90d4820bace0 & $0.2902$ & $0.2937$ & $\mathbf{0.2558}$ & $0.2749$ & $0.3323$ \\
bad\_customer\_detection-7b2ea3462dce & $\mathbf{0.2474}$ & $0.2509$ & $0.2536$ & $0.2609$ & $0.2496$ \\
bank\_customer\_churn-b1fc11f3ea36 & $0.1250$ & $\mathbf{0.1236}$ & $0.1290$ & $0.1287$ & $0.1290$ \\
bank\_marketing-ba2b45409a51 & $0.2349$ & $\mathbf{0.2336}$ & $0.2348$ & $0.2377$ & $0.2351$ \\
biogeographical\_ancestry\_prediction- & $\mathbf{1.3388}$ & $1.3726$ & $1.3413$ & $1.4056$ & $1.4878$ \\
biomechanical\_orthopaedic\_prediction & $0.3497$ & $\mathbf{0.3495}$ & $0.3511$ & $0.3522$ & $0.3552$ \\
bioresponse-63f85b558d70 & $0.1293$ & $0.1263$ & $0.1437$ & $\mathbf{0.1259}$ & $0.1307$ \\
blood\_transfusion-c0b8fbb104f3 & $0.2501$ & $\mathbf{0.2496}$ & $0.2528$ & $0.2559$ & $0.2599$ \\
cardiotocography-172ef52f9a1b & $0.2188$ & $0.2490$ & $0.2343$ & $0.2288$ & $\mathbf{0.2146}$ \\
churn-4ae7db677764 & $\mathbf{0.0618}$ & $0.0696$ & $0.0632$ & $0.0785$ & $0.0776$ \\
clock\_protein\_toxicity-f1eed69eb784 & $\mathbf{0.4823}$ & $0.4989$ & $0.5169$ & $0.4858$ & $0.4915$ \\
coil\_2000-d498c6d7849f & $\mathbf{0.2191}$ & $0.2200$ & $0.2199$ & $0.2364$ & $0.2330$ \\
consumer\_complaints\_1m-853af219eb5a & $0.5072$ & $\mathbf{0.4104}$ & $0.4452$ & $0.4353$ & $0.4251$ \\
covertype-1ed1c3d23864 & $1.0588$ & $0.9675$ & $1.0253$ & $\mathbf{0.7115}$ & $0.7172$ \\
credit\_approval-a37d9105992b & $\mathbf{0.0622}$ & $0.0665$ & $0.0638$ & $0.0682$ & $0.0670$ \\
credit\_card\_clients\_default-9470ea9 & $\mathbf{0.2112}$ & $0.2126$ & $0.2116$ & $0.2175$ & $0.2191$ \\
credit\_g-6ced9fa4ef3f & $\mathbf{0.1933}$ & $0.2005$ & $0.2008$ & $0.2025$ & $0.2008$ \\
customer\_satisfaction\_in\_airline-71 & $0.0051$ & $\mathbf{0.0046}$ & $0.0048$ & $0.0050$ & $0.0053$ \\
dementia\_prediction-d8ce15bd8d63 & $0.6785$ & $0.7190$ & $0.6937$ & $\mathbf{0.6136}$ & $0.6355$ \\
diabetes\_130\_us-a06cb03600f0 & $0.3355$ & $0.3331$ & $0.3392$ & $0.3321$ & $\mathbf{0.3290}$ \\
drug\_induced\_autoimmunity\_predictio & $\mathbf{0.1445}$ & $0.1630$ & $0.1519$ & $0.1739$ & $0.1811$ \\
early\_stage\_diabetes\_risk\_predicti & $\mathbf{0.0235}$ & $0.0273$ & $0.0241$ & $0.0366$ & $0.0243$ \\
ecoli\_proteins-88ead48fb030 & $\mathbf{0.3512}$ & $0.3626$ & $0.3573$ & $0.3613$ & $0.3666$ \\
ecommerce\_shipping-db33e797abb9 & $\mathbf{0.2524}$ & $0.2534$ & $0.2549$ & $0.2551$ & $0.2562$ \\
emscad-1790bb44ad91 & $\mathbf{0.0047}$ & $0.0081$ & $0.0060$ & $0.0073$ & $0.0120$ \\
eryhemato\_squamous\_disease-a64cf6ea9 & $\mathbf{0.0448}$ & $0.0612$ & $0.0543$ & $0.0698$ & $0.0661$ \\
fitness\_club-95daf6614f7a & $0.1799$ & $\mathbf{0.1792}$ & $0.1795$ & $0.1818$ & $0.1853$ \\
forensic\_glass\_identification-e30419 & $\mathbf{0.6417}$ & $0.6875$ & $0.6895$ & $0.7718$ & $0.7654$ \\
gallstone\_disease-63982cfe3e7c & $0.0907$ & $\mathbf{0.0798}$ & $0.0858$ & $0.1086$ & $0.1395$ \\
ghanas\_indigenous\_intel-ecbbda50d44e & $0.1839$ & $0.1988$ & $0.2428$ & $\mathbf{0.1739}$ & $0.1893$ \\
give\_me\_some\_credit-51f4a5f15bd6 & $0.1516$ & $\mathbf{0.1363}$ & $0.1511$ & $0.1518$ & $0.1537$ \\
hazelnut\_spread\_contaminant\_detecti & $0.0027$ & $0.0036$ & $\mathbf{0.0025}$ & $0.0141$ & $0.0274$ \\
heart\_disease\_cleveland-1e633334e0ef & $0.0960$ & $0.0935$ & $\mathbf{0.0935}$ & $0.0947$ & $0.0943$ \\
heart\_disease\_hungary-f5ca5cfb8352 & $0.1019$ & $\mathbf{0.0925}$ & $0.0983$ & $0.0975$ & $0.1075$ \\
heart\_disease\_va\_long\_beach-062b5d & $0.3227$ & $0.3265$ & $0.3301$ & $0.3573$ & $\mathbf{0.3183}$ \\
heart\_failure\_followup\_survival-25d & $\mathbf{0.0784}$ & $0.0921$ & $0.0851$ & $0.0958$ & $0.0893$ \\
heloc-d2ac12c6b994 & $\mathbf{0.2072}$ & $0.2074$ & $0.2074$ & $0.2077$ & $0.2109$ \\
hepatitis\_c\_prediction-d17a9e02b43b & $\mathbf{0.0998}$ & $0.1048$ & $0.1156$ & $0.1237$ & $0.1425$ \\
hepatitis\_survival\_prediction-d16737 & $\mathbf{0.1181}$ & $0.1358$ & $0.1272$ & $0.1442$ & $0.1331$ \\
hiva\_agnostic-c4b35582c204 & $\mathbf{0.2102}$ & $0.2171$ & $0.2491$ & $0.2211$ & $0.2164$ \\
home\_credit\_default\_risk-1b1ae7eb05 & $0.2347$ & $0.2134$ & $0.2232$ & $\mathbf{0.2076}$ & $0.2079$ \\
home\_credit\_default\_stability\_1m-e & $0.1593$$^{\dagger}$ & $0.1442$ & $0.1639$ & $\mathbf{0.1279}$ & $0.1310$ \\
homeq\_default\_prediction-6da1464558b & $\mathbf{0.0000}$ & $0.0001$ & $0.0012$ & $0.0074$ & $0.0220$ \\
homesite\_quote\_conversion-f57214c4e0 & $0.0360$ & $0.0360$ & $0.0413$ & $0.0322$ & $\mathbf{0.0318}$ \\
horse\_colic\_survival-e1beb13d0b4e & $\mathbf{0.6813}$ & $0.7029$ & $0.7058$ & $0.6997$ & $0.7282$ \\
hotel\_booking\_demand-9541da6f5a2b & $0.2073$ & $0.2289$ & $0.2439$ & $0.1826$ & $\mathbf{0.1703}$ \\
hr\_analytics-9f0cb22a2cd4 & $\mathbf{0.1952}$ & $0.1960$ & $0.1958$ & $0.1978$ & $0.1966$ \\
ieee\_fraud\_detection-5e0af5cbbb73 & $0.1212$ & $0.1008$ & $0.1096$ & $\mathbf{0.0827}$ & $0.0870$ \\
in\_vehicle\_coupon\_recommendation-10 & $\mathbf{0.1529}$ & $0.1583$ & $0.1619$ & $0.1558$ & $0.1577$ \\
indian\_liver\_patient\_dataset-0216cd & $0.2313$ & $0.2362$ & $\mathbf{0.2285}$ & $0.2465$ & $0.2475$ \\
iranian\_churn-87d6fa636729 & $\mathbf{0.0026}$ & $0.0027$ & $0.0030$ & $0.0046$ & $0.0096$ \\
jm1-c8cfe392b650 & $\mathbf{0.2238}$ & $0.2290$ & $0.2309$ & $0.2495$ & $0.2468$ \\
kdd\_cup\_09\_appetency-72d0143e7c3d & $0.1555$ & $0.1649$ & $0.1556$ & $0.1732$ & $\mathbf{0.1428}$ \\
kick-57821d387256 & $\mathbf{0.2049}$ & $0.2049$ & $0.2058$ & $0.2070$ & $0.2076$ \\
kickstarter-fc5123d788a6 & $0.0755$ & $0.0694$ & $0.0765$ & $\mathbf{0.0679}$ & $0.0685$ \\
labour\_inspection\_compliance-17b501f & $0.2448$ & $\mathbf{0.2083}$ & $0.2084$ & $0.2105$ & $0.2156$ \\
lending\_club\_1m-f6bf13e264e4 & $0.3047$$^{\dagger}$ & $0.2962$ & $0.3148$ & $0.2851$ & $\mathbf{0.2850}$ \\
ljubljana\_breast\_cancer-28c8b88c6b23 & $\mathbf{0.2732}$ & $0.2789$ & $0.2782$ & $0.3106$ & $0.3046$ \\
ljubljana\_primary\_tumor-b1acf2d83b0d & $\mathbf{1.3092}$ & $1.3216$ & $1.3328$ & $1.3175$ & $1.3763$ \\
lung\_cancer-48c9b1f8934d & $0.2225$ & $\mathbf{0.1269}$ & $0.1913$ & $0.1459$ & $0.2041$ \\
lung\_cancer\_epithelial\_genexp-ff623 & $\mathbf{0.2025}$ & $0.2426$ & $0.2773$ & $0.2482$ & $0.2536$ \\
marketing\_campaign-45b76b0f0de6 & $0.0869$ & $0.0936$ & $\mathbf{0.0869}$ & $0.1021$ & $0.1083$ \\
maternal\_health\_risk-002da4815804 & $0.3027$ & $\mathbf{0.2971}$ & $0.2994$ & $0.3567$ & $0.4010$ \\
mic-ba10980b94f8 & $\mathbf{0.4275}$ & $0.4424$ & $0.4416$ & $0.4375$ & $0.4642$ \\
mice\_protein\_trisomy\_discriminant-1 & $\mathbf{0.7885}$ & $1.1642$ & $0.9219$ & $1.2192$ & $1.3395$ \\
micro\_mass-5deb16d4fa0e & $\mathbf{0.3948}$ & $0.6937$ & $0.5250$ & $0.6203$ & $0.5135$ \\
musk-16dca0dbddf5 & $0.1567$ & $0.1281$ & $\mathbf{0.1090}$ & $0.1226$ & $0.1926$ \\
naticusdroid\_android\_permissions\_da & $\mathbf{0.0117}$ & $0.0125$ & $0.0123$ & $0.0132$ & $0.0130$ \\
online\_shoppers\_purchasing\_intentio & $0.0559$ & $\mathbf{0.0556}$ & $0.0568$ & $0.0606$ & $0.0607$ \\
otto\_group\_product\_classification\_ & $0.4681$ & $0.4392$ & $\mathbf{0.4175}$ & $0.4374$ & $0.4611$ \\
pancreatic\_cancer\_mouse\_detection-b & $\mathbf{0.3255}$ & $0.3645$ & $0.4940$ & $0.4442$ & $0.4648$ \\
parkinsons\_biomedical\_voice\_measure & $0.1902$ & $0.2432$ & $0.1700$ & $\mathbf{0.1256}$ & $0.1507$ \\
polish\_companies\_bankruptcy-c6e7be38 & $\mathbf{0.0041}$ & $0.0090$ & $0.0105$ & $0.0291$ & $0.0310$ \\
porto\_seguro-7894967113f3 & $0.3638$ & $0.3582$ & $0.3654$ & $0.3575$ & $\mathbf{0.3571}$ \\
predict\_students\_dropout\_and\_acade & $\mathbf{0.5206}$ & $0.5358$ & $0.5287$ & $0.5341$ & $0.5369$ \\
prostate\_cancer\_detection-d59b58af23 & $\mathbf{0.0038}$ & $0.0072$ & $0.0606$ & $0.0171$ & $0.0179$ \\
pva\_revenue\_prediction\_kddcup98-94a & $0.4218$ & $0.3838$ & $0.3951$ & $0.3731$ & $\mathbf{0.3689}$ \\
qsar\_biodeg-99edbe20f66b & $\mathbf{0.0564}$ & $0.0577$ & $0.0564$ & $0.0615$ & $0.0641$ \\
regensburg\_pediatric\_appendicitis-e7 & $0.0634$ & $\mathbf{0.0628}$ & $0.0644$ & $0.0772$ & $0.0776$ \\
santander\_customer\_satisfaction-5042 & $\mathbf{0.1513}$ & $0.1644$ & $0.1862$ & $0.1547$ & $0.1554$ \\
santander\_customer\_transaction\_pred & $0.1247$ & $0.1183$ & $0.1157$ & $0.1021$ & $\mathbf{0.0921}$ \\
sdss\_17-3d9b16fdea21 & $0.0686$ & $\mathbf{0.0601}$ & $0.0666$ & $0.0835$ & $0.0708$ \\
seismic\_bumps-9636045db12f & $\mathbf{0.2080}$ & $0.2115$ & $0.2146$ & $0.2265$ & $0.2194$ \\
sepsis\_prediction\_1m-77a236a7d8eb & $\mathbf{0.1820}$ & $0.2255$ & $0.2013$ & $0.1821$ & $0.1877$ \\
sepsis\_survival\_minimal\_clinical\_r & $0.2933$ & $0.2933$ & $0.2980$ & $\mathbf{0.2932}$ & $0.2933$ \\
south\_africa\_coronary\_heart\_diseas & $0.2310$ & $0.2281$ & $\mathbf{0.2276}$ & $0.2394$ & $0.2437$ \\
splice-0b61e96684a9 & $\mathbf{0.1094}$ & $0.1167$ & $0.1137$ & $0.1244$ & $0.1283$ \\
taiwanese\_bankruptcy\_prediction-e1d1 & $\mathbf{0.0483}$ & $0.0484$ & $0.0509$ & $0.0523$ & $0.0531$ \\
thyroid\_discordant-1111dc4e4f7d & $0.0111$ & $0.0092$ & $\mathbf{0.0091}$ & $0.0126$ & $0.0204$ \\
tour\_travels\_churn-638a10e1adbe & $0.0395$ & $\mathbf{0.0380}$ & $0.0417$ & $0.0447$ & $0.0420$ \\
website\_phishing-e1ffdd9e6d3a & $0.2555$ & $\mathbf{0.2443}$ & $0.2500$ & $0.2654$ & $0.2734$ \\
wids\_diabetes\_mellitus-d05f898265f2 & $0.1379$ & $0.1415$ & $0.1414$ & $\mathbf{0.1216}$ & $0.1260$ \\
\midrule
Mean & $0.2545$ & $0.2621$ & $0.2620$ & $0.2617$ & $0.2672$ \\
Mean rank & $2.28$ & $2.68$ & $3.06$ & $3.29$ & $3.69$ \\
\end{longtable}
}

{\scriptsize
\setlength{\tabcolsep}{3pt}
\begin{longtable}{@{}lrrrrrrrrrr@{}}
\caption{Fit and prediction times in seconds, averaged over splits. FTFM uses
one H100, while baselines use the benchmark maintainers' hardware, so absolute times
are not directly comparable. \Cref{fig:predict_scaling} compares scaling with
table size, which a constant hardware factor preserves. $^{\dagger}$ marks
FTFM's two reduced-support results (rung~$3$).}
\label{tab:beyondarena_cost} \\
\toprule
& \multicolumn{2}{c}{FTFM} & \multicolumn{2}{c}{TabPFN-3} & \multicolumn{2}{c}{TabICLv2} & \multicolumn{2}{c}{RealMLP} & \multicolumn{2}{c}{CatBoost} \\
\cmidrule(lr){2-3}\cmidrule(lr){4-5}\cmidrule(lr){6-7}\cmidrule(lr){8-9}\cmidrule(lr){10-11}
Dataset & fit & pred. & fit & pred. & fit & pred. & fit & pred. & fit & pred. \\
\midrule
\endfirsthead
\toprule
& \multicolumn{2}{c}{FTFM} & \multicolumn{2}{c}{TabPFN-3} & \multicolumn{2}{c}{TabICLv2} & \multicolumn{2}{c}{RealMLP} & \multicolumn{2}{c}{CatBoost} \\
\cmidrule(lr){2-3}\cmidrule(lr){4-5}\cmidrule(lr){6-7}\cmidrule(lr){8-9}\cmidrule(lr){10-11}
Dataset & fit & pred. & fit & pred. & fit & pred. & fit & pred. & fit & pred. \\
\midrule
\endhead
\midrule
\multicolumn{11}{r}{\emph{continued on the next page}} \\
\endfoot
\bottomrule
\endlastfoot
acquire\_valued\_shoppers\_challenge-f & $3643$ & $51.0$ & $454$ & $59.9$ & $344$ & $35.4$ & $80328$ & $19.5$ & $5064$ & $2.2$ \\
amazon\_employee\_access-4417c4c952bc & $208$ & $5.8$ & $45$ & $4.7$ & $20$ & $2.3$ & $9594$ & $24.5$ & $1652$ & $6.3$ \\
amex\_non\_iid\_1m-e2e64ee8b42e & $20596$ & $1320.8$ & $20049$ & $5953.4$ & $5329$ & $839.0$ & $278841$ & $766.3$ & $85907$ & $109.0$ \\
anes\_voting\_2026-58a0e941922e & $2592$ & $38.7$ & $320$ & $23.3$ & $360$ & $27.4$ & $90522$ & $134.7$ & $10293$ & $29.0$ \\
aps\_failure-b63fc6330117 & $3203$ & $137.7$ & $204$ & $31.2$ & $229$ & $22.1$ & $49991$ & $141.5$ & $4118$ & $1.9$ \\
asp\_potassco\_classification-6cbac33e & $352$ & $13.4$ & $22$ & $1.1$ & $26$ & $1.5$ & $2137$ & $40.2$ & $2584$ & $0.5$ \\
audiology\_diagnosis-90d4820bace0 & $63$ & $0.4$ & $57$ & $1.1$ & $25$ & $0.6$ & $3718$ & $20.4$ & $737$ & $1.1$ \\
bad\_customer\_detection-7b2ea3462dce & $25$ & $0.5$ & $25$ & $1.1$ & $14$ & $1.0$ & $1659$ & $18.1$ & $274$ & $1.1$ \\
bank\_customer\_churn-b1fc11f3ea36 & $71$ & $1.5$ & $31$ & $1.7$ & $13$ & $0.9$ & $3749$ & $11.6$ & $352$ & $0.4$ \\
bank\_marketing-ba2b45409a51 & $329$ & $10.7$ & $62$ & $7.1$ & $28$ & $3.5$ & $15964$ & $23.1$ & $697$ & $2.3$ \\
biogeographical\_ancestry\_prediction- & $96$ & $0.8$ & $81$ & $1.5$ & $48$ & $0.7$ & $3411$ & $42.8$ & $4115$ & $10.0$ \\
biomechanical\_orthopaedic\_prediction & $65$ & $0.3$ & $51$ & $1.0$ & $21$ & $0.5$ & $3449$ & $8.5$ & $1004$ & $0.3$ \\
bioresponse-63f85b558d70 & $875$ & $33.3$ & $125$ & $6.6$ & $205$ & $11.8$ & $10586$ & $344.2$ & $3981$ & $7.5$ \\
blood\_transfusion-c0b8fbb104f3 & $68$ & $0.4$ & $52$ & $1.1$ & $21$ & $0.6$ & $3259$ & $18.1$ & $507$ & $0.3$ \\
cardiotocography-172ef52f9a1b & $32$ & $0.6$ & $25$ & $1.5$ & $12$ & $0.6$ & $1881$ & $8.8$ & $443$ & $0.5$ \\
churn-4ae7db677764 & $49$ & $1.0$ & $32$ & $1.5$ & $11$ & $0.7$ & $2927$ & $15.4$ & $561$ & $0.6$ \\
clock\_protein\_toxicity-f1eed69eb784 & $215$ & $1.2$ & $91$ & $1.4$ & $128$ & $0.9$ & $6919$ & $293.1$ & $4061$ & $4.4$ \\
coil\_2000-d498c6d7849f & $150$ & $4.8$ & $46$ & $3.3$ & $24$ & $1.6$ & $4890$ & $21.9$ & $404$ & $1.7$ \\
consumer\_complaints\_1m-853af219eb5a & $7501$ & $1045.4$ & $19444$ & $3250.3$ & $23070$ & $3766.8$ & $304315$ & $297.7$ & $322811$ & $110.3$ \\
covertype-1ed1c3d23864 & $3065$ & $661.6$ & $381$ & $486.6$ & $416$ & $359.1$ & $37167$ & $8.2$ & $648$ & $1.5$ \\
credit\_approval-a37d9105992b & $71$ & $0.4$ & $53$ & $1.1$ & $23$ & $0.6$ & $3641$ & $31.7$ & $588$ & $1.6$ \\
credit\_card\_clients\_default-9470ea9 & $236$ & $7.0$ & $40$ & $4.3$ & $24$ & $2.4$ & $9202$ & $17.4$ & $491$ & $1.1$ \\
credit\_g-6ced9fa4ef3f & $28$ & $0.4$ & $22$ & $1.1$ & $12$ & $0.7$ & $1555$ & $19.4$ & $303$ & $1.3$ \\
customer\_satisfaction\_in\_airline-71 & $1381$ & $65.1$ & $207$ & $37.0$ & $119$ & $18.4$ & $45612$ & $28.6$ & $3820$ & $4.3$ \\
dementia\_prediction-d8ce15bd8d63 & $59$ & $0.3$ & $48$ & $1.0$ & $21$ & $0.6$ & $3421$ & $14.1$ & $673$ & $0.2$ \\
diabetes\_130\_us-a06cb03600f0 & $766$ & $36.3$ & $122$ & $16.0$ & $73$ & $8.9$ & $20359$ & $44.5$ & $2813$ & $8.6$ \\
drug\_induced\_autoimmunity\_predictio & $112$ & $0.9$ & $58$ & $1.2$ & $41$ & $0.7$ & $4068$ & $60.9$ & $2027$ & $1.2$ \\
early\_stage\_diabetes\_risk\_predicti & $59$ & $0.4$ & $51$ & $1.1$ & $21$ & $0.6$ & $3488$ & $19.7$ & $901$ & $0.6$ \\
ecoli\_proteins-88ead48fb030 & $64$ & $0.3$ & $46$ & $1.0$ & $22$ & $0.6$ & $3637$ & $15.7$ & $659$ & $0.3$ \\
ecommerce\_shipping-db33e797abb9 & $72$ & $1.6$ & $26$ & $1.7$ & $12$ & $0.9$ & $4308$ & $18.8$ & $287$ & $0.4$ \\
emscad-1790bb44ad91 & $614$ & $26.1$ & $101$ & $11.8$ & $99$ & $11.2$ & $14921$ & $78.3$ & $6320$ & $10.3$ \\
eryhemato\_squamous\_disease-a64cf6ea9 & $69$ & $0.4$ & $62$ & $1.2$ & $30$ & $0.6$ & $3479$ & $5.8$ & $6047$ & $3.0$ \\
fitness\_club-95daf6614f7a & $29$ & $0.4$ & $23$ & $1.1$ & $9$ & $0.7$ & $1587$ & $12.2$ & $256$ & $0.4$ \\
forensic\_glass\_identification-e30419 & $60$ & $0.4$ & $45$ & $1.0$ & $20$ & $0.5$ & $3490$ & $10.8$ & $1387$ & $0.4$ \\
gallstone\_disease-63982cfe3e7c & $63$ & $0.3$ & $49$ & $1.1$ & $25$ & $0.6$ & $3475$ & $34.9$ & $813$ & $0.8$ \\
ghanas\_indigenous\_intel-ecbbda50d44e & $85$ & $1.3$ & $31$ & $1.8$ & $23$ & $1.1$ & $4560$ & $12.3$ & $2506$ & $0.7$ \\
give\_me\_some\_credit-51f4a5f15bd6 & $1526$ & $68.1$ & $253$ & $46.2$ & $125$ & $22.0$ & $54981$ & $35.5$ & $1175$ & $2.9$ \\
hazelnut\_spread\_contaminant\_detecti & $38$ & $0.7$ & $19$ & $1.6$ & $13$ & $0.7$ & $2160$ & $10.3$ & $1093$ & $0.9$ \\
heart\_disease\_cleveland-1e633334e0ef & $63$ & $0.3$ & $50$ & $1.0$ & $22$ & $0.6$ & $3610$ & $23.5$ & $648$ & $1.2$ \\
heart\_disease\_hungary-f5ca5cfb8352 & $67$ & $0.3$ & $53$ & $1.1$ & $23$ & $0.6$ & $3505$ & $24.8$ & $621$ & $1.4$ \\
heart\_disease\_va\_long\_beach-062b5d & $63$ & $0.3$ & $49$ & $1.1$ & $22$ & $0.6$ & $3469$ & $30.0$ & $597$ & $1.6$ \\
heart\_failure\_followup\_survival-25d & $68$ & $0.3$ & $48$ & $1.1$ & $22$ & $0.6$ & $3437$ & $28.6$ & $622$ & $0.6$ \\
heloc-d2ac12c6b994 & $87$ & $2.2$ & $26$ & $1.9$ & $15$ & $1.0$ & $3669$ & $12.2$ & $363$ & $0.6$ \\
hepatitis\_c\_prediction-d17a9e02b43b & $68$ & $0.4$ & $47$ & $1.1$ & $22$ & $0.6$ & $3490$ & $13.2$ & $1006$ & $0.3$ \\
hepatitis\_survival\_prediction-d16737 & $61$ & $0.3$ & $50$ & $1.1$ & $23$ & $0.6$ & $3519$ & $33.7$ & $578$ & $1.0$ \\
hiva\_agnostic-c4b35582c204 & $708$ & $27.5$ & $125$ & $6.5$ & $165$ & $10.9$ & $3751$ & $544.8$ & $1882$ & $4.5$ \\
home\_credit\_default\_risk-1b1ae7eb05 & $11501$ & $199.4$ & $1992$ & $267.0$ & $12877$ & $2688.5$ & $236089$ & $858.6$ & $86390$ & $70.1$ \\
home\_credit\_default\_stability\_1m-e & $38863$ & $714.7$$^{\dagger}$ & $24019$ & $5980.7$ & $6072$ & $854.0$ & $321564$ & $2054.3$ & $239805$ & $125.9$ \\
homeq\_default\_prediction-6da1464558b & $57$ & $1.2$ & $22$ & $1.4$ & $10$ & $0.7$ & $3694$ & $13.0$ & $502$ & $0.8$ \\
homesite\_quote\_conversion-f57214c4e0 & $19087$ & $79.7$ & $1363$ & $201.3$ & $2017$ & $286.9$ & $203535$ & $291.9$ & $43788$ & $32.8$ \\
horse\_colic\_survival-e1beb13d0b4e & $68$ & $0.3$ & $52$ & $1.2$ & $24$ & $0.6$ & $3363$ & $20.8$ & $487$ & $1.6$ \\
hotel\_booking\_demand-9541da6f5a2b & $1128$ & $15.1$ & $169$ & $18.4$ & $108$ & $9.9$ & $34747$ & $23.6$ & $5955$ & $2.9$ \\
hr\_analytics-9f0cb22a2cd4 & $123$ & $3.1$ & $31$ & $2.7$ & $17$ & $1.4$ & $5999$ & $19.1$ & $439$ & $1.7$ \\
ieee\_fraud\_detection-5e0af5cbbb73 & $19714$ & $196.4$ & $5279$ & $680.9$ & $30517$ & $4448.8$ & $282144$ & $880.1$ & $175089$ & $190.0$ \\
in\_vehicle\_coupon\_recommendation-10 & $108$ & $2.8$ & $31$ & $2.2$ & $16$ & $1.2$ & $5794$ & $26.3$ & $1027$ & $2.8$ \\
indian\_liver\_patient\_dataset-0216cd & $64$ & $0.4$ & $47$ & $1.0$ & $21$ & $0.6$ & $3438$ & $32.7$ & $462$ & $0.3$ \\
iranian\_churn-87d6fa636729 & $39$ & $0.6$ & $20$ & $1.4$ & $12$ & $0.6$ & $2445$ & $8.1$ & $342$ & $0.5$ \\
jm1-c8cfe392b650 & $97$ & $2.4$ & $25$ & $1.9$ & $15$ & $1.0$ & $5162$ & $20.0$ & $602$ & $0.6$ \\
kdd\_cup\_09\_appetency-72d0143e7c3d & $2268$ & $98.3$ & $192$ & $21.9$ & $205$ & $19.7$ & $22486$ & $171.6$ & $4082$ & $10.2$ \\
kick-57821d387256 & $1155$ & $13.2$ & $163$ & $17.6$ & $97$ & $9.4$ & $28331$ & $20.0$ & $4596$ & $5.7$ \\
kickstarter-fc5123d788a6 & $6414$ & $83.9$ & $734$ & $93.5$ & $671$ & $61.6$ & $98659$ & $17.6$ & $44249$ & $8.1$ \\
labour\_inspection\_compliance-17b501f & $6227$ & $269.1$ & $295$ & $32.8$ & $448$ & $46.3$ & $71565$ & $410.6$ & $10763$ & $22.7$ \\
lending\_club\_1m-f6bf13e264e4 & $9278$ & $358.9$$^{\dagger}$ & $14856$ & $2447.1$ & $5934$ & $641.4$ & $301036$ & $458.5$ & $149009$ & $111.4$ \\
ljubljana\_breast\_cancer-28c8b88c6b23 & $65$ & $0.3$ & $50$ & $1.1$ & $21$ & $0.6$ & $3665$ & $28.7$ & $431$ & $1.0$ \\
ljubljana\_primary\_tumor-b1acf2d83b0d & $198$ & $1.1$ & $51$ & $1.1$ & $32$ & $0.9$ & $3536$ & $27.9$ & $746$ & $0.8$ \\
lung\_cancer-48c9b1f8934d & $2971$ & $12.9$ & $5727$ & $7.7$ & $2511$ & $10.3$ & $37525$ & $14.9$ & $205221$ & $4.9$ \\
lung\_cancer\_epithelial\_genexp-ff623 & $7174$ & $23.5$ & $19316$ & $14.2$ & $7570$ & $22.2$ & $75574$ & $734.1$ & $187324$ & $25.2$ \\
marketing\_campaign-45b76b0f0de6 & $36$ & $0.7$ & $29$ & $1.7$ & $13$ & $0.8$ & $1927$ & $20.3$ & $327$ & $1.1$ \\
maternal\_health\_risk-002da4815804 & $27$ & $0.4$ & $19$ & $0.9$ & $11$ & $0.6$ & $1498$ & $7.4$ & $310$ & $0.4$ \\
mic-ba10980b94f8 & $55$ & $1.4$ & $33$ & $1.6$ & $21$ & $0.9$ & $1753$ & $33.1$ & $2984$ & $2.8$ \\
mice\_protein\_trisomy\_discriminant-1 & $132$ & $1.1$ & $52$ & $1.1$ & $39$ & $0.8$ & $4279$ & $25.6$ & $12878$ & $0.7$ \\
micro\_mass-5deb16d4fa0e & $2188$ & $24.2$ & $104$ & $1.4$ & $173$ & $2.4$ & $10876$ & $13.2$ & $105052$ & $5.5$ \\
musk-16dca0dbddf5 & $564$ & $7.2$ & $159$ & $5.0$ & $90$ & $2.4$ & $12441$ & $29.3$ & $6711$ & $1.0$ \\
naticusdroid\_android\_permissions\_da & $118$ & $3.8$ & $40$ & $3.2$ & $21$ & $1.4$ & $3143$ & $33.8$ & $530$ & $0.9$ \\
online\_shoppers\_purchasing\_intentio & $90$ & $2.2$ & $37$ & $2.0$ & $13$ & $1.1$ & $4610$ & $17.4$ & $436$ & $1.4$ \\
otto\_group\_product\_classification\_ & $1011$ & $40.7$ & $109$ & $16.5$ & $99$ & $10.5$ & $28086$ & $34.7$ & $30348$ & $4.2$ \\
pancreatic\_cancer\_mouse\_detection-b & $1264$ & $6.6$ & $1778$ & $5.0$ & $946$ & $3.9$ & $15044$ & $600.7$ & $34315$ & $9.3$ \\
parkinsons\_biomedical\_voice\_measure & $63$ & $0.4$ & $50$ & $1.1$ & $28$ & $0.6$ & $3780$ & $22.2$ & $1958$ & $0.5$ \\
polish\_companies\_bankruptcy-c6e7be38 & $103$ & $3.1$ & $30$ & $2.2$ & $15$ & $1.0$ & $3475$ & $26.4$ & $872$ & $0.5$ \\
porto\_seguro-7894967113f3 & $1976$ & $1128.7$ & $3208$ & $651.5$ & $1607$ & $308.4$ & $194355$ & $128.8$ & $11322$ & $32.1$ \\
predict\_students\_dropout\_and\_acade & $53$ & $1.3$ & $28$ & $1.8$ & $16$ & $0.8$ & $2513$ & $14.9$ & $677$ & $1.7$ \\
prostate\_cancer\_detection-d59b58af23 & $4460$ & $26.0$ & $8025$ & $9.1$ & $3409$ & $19.7$ & $42337$ & $426.2$ & $198639$ & $14.8$ \\
pva\_revenue\_prediction\_kddcup98-94a & $11834$ & $182.6$ & $1270$ & $122.8$ & $20451$ & $4257.7$ & $233580$ & $879.8$ & $32685$ & $80.9$ \\
qsar\_biodeg-99edbe20f66b & $27$ & $0.5$ & $21$ & $1.1$ & $14$ & $0.7$ & $1584$ & $11.3$ & $384$ & $0.7$ \\
regensburg\_pediatric\_appendicitis-e7 & $38$ & $1.0$ & $27$ & $1.3$ & $18$ & $0.8$ & $1814$ & $38.7$ & $358$ & $2.3$ \\
santander\_customer\_satisfaction-5042 & $2318$ & $97.8$ & $222$ & $31.6$ & $272$ & $26.6$ & $42495$ & $66.9$ & $2524$ & $2.4$ \\
santander\_customer\_transaction\_pred & $21543$ & $88.3$ & $1011$ & $139.3$ & $2910$ & $445.9$ & $123073$ & $675.0$ & $63677$ & $76.0$ \\
sdss\_17-3d9b16fdea21 & $607$ & $23.2$ & $95$ & $15.5$ & $49$ & $7.6$ & $18834$ & $9.2$ & $8615$ & $10.0$ \\
seismic\_bumps-9636045db12f & $32$ & $0.6$ & $21$ & $1.4$ & $12$ & $0.7$ & $1820$ & $16.2$ & $263$ & $0.6$ \\
sepsis\_prediction\_1m-77a236a7d8eb & $26806$ & $999.5$ & $17013$ & $3021.8$ & $9655$ & $1663.1$ & $286563$ & $271.5$ & $7643$ & $16.6$ \\
sepsis\_survival\_minimal\_clinical\_r & $854$ & $35.3$ & $155$ & $27.7$ & $67$ & $12.0$ & $35064$ & $22.9$ & $406$ & $0.4$ \\
south\_africa\_coronary\_heart\_diseas & $61$ & $0.3$ & $45$ & $1.0$ & $21$ & $0.6$ & $3655$ & $28.7$ & $521$ & $0.4$ \\
splice-0b61e96684a9 & $53$ & $1.5$ & $38$ & $2.1$ & $20$ & $0.8$ & $2326$ & $22.9$ & $1473$ & $3.4$ \\
taiwanese\_bankruptcy\_prediction-e1d1 & $116$ & $3.5$ & $38$ & $2.8$ & $22$ & $1.3$ & $3707$ & $18.5$ & $823$ & $0.8$ \\
thyroid\_discordant-1111dc4e4f7d & $44$ & $1.0$ & $29$ & $1.6$ & $12$ & $0.7$ & $2008$ & $13.7$ & $454$ & $0.9$ \\
tour\_travels\_churn-638a10e1adbe & $26$ & $0.4$ & $20$ & $1.1$ & $11$ & $0.6$ & $2084$ & $11.8$ & $281$ & $0.3$ \\
website\_phishing-e1ffdd9e6d3a & $25$ & $0.4$ & $21$ & $1.1$ & $14$ & $0.7$ & $1984$ & $7.7$ & $256$ & $0.6$ \\
wids\_diabetes\_mellitus-d05f898265f2 & $6447$ & $279.0$ & $506$ & $72.0$ & $489$ & $51.6$ & $65939$ & $212.3$ & $20996$ & $13.6$ \\
\midrule
Median & $110$ & $2.2$ & $51$ & $1.9$ & $24$ & $1.0$ & $3924$ & $24.0$ & $1016$ & $1.5$ \\
\end{longtable}
}

\section{Serving policy on BeyondArena}
\label{app:serving_policy}

BeyondArena fixes datasets, splits, preprocessing, bagging, and metrics.
Our serving settings follow \cref{tab:serving_ladder}, and we record the lowest
feasible rung for each dataset. Table shape determines rungs~$0$ and~$1$.
Rungs~$2$ and~$3$ apply only after a failure at the preceding rung. The released
code identifies the affected datasets.

\begin{table}[h]
\centering
\small
\caption{Serving settings for the $98$ BeyondArena classification datasets.
$T$ and $M$ count support rows and columns after preprocessing, with
$\mathrm{work}=T^2M$ and $\mathrm{cells}=TM$. All rungs use fp32, including
attention. Higher rungs reduce ensemble size, increase the time budget, then
reduce support.}
\label{tab:serving_ladder}
\begin{tabular}{@{}clllllr@{}}
\toprule
Rung & Serving & Attention & Members & Budget & Support & $n$ \\
\midrule
$0$ & fp32 & fp32 & $8$ & $4$\,h & full & $88$ \\
$1$ & fp32 & fp32 & $1$ & $30{,}600$\,s & full & $2$ \\
$2$ & fp32 & fp32 & $1$ & $80{,}000$\,s & full & $6$ \\
$3$ & fp32 & fp32 & $1$ & $80{,}000$\,s & $500$k rows & $2$ \\
\bottomrule
\end{tabular}
\end{table}

\paragraph{Precision.} We retain fp32 for consistent numerics. FlashAttention
has no fp32 kernel, so we use the memory-efficient kernel, whose $T^2$ cost
can exhaust a fold's budget. The largest completed table had
$\mathrm{work}=6.894\times10^{11}$. The smallest table that failed had
$8.999\times10^{11}$ and took $2{,}474$\,s for one fold.

\paragraph{Escalation.} Rung~$1$ applies when
$\mathrm{work}>1.2\times10^{13}$ or $\mathrm{cells}>4.3\times10^{7}$. The first
threshold comes from the largest table that completed with eight members
($744$\,s per fold). Above the second, the cell grid exceeds accelerator memory
and prefill streams from host memory, about $4.5\times$ slower. Reducing the
ensemble costs about $1\%$ in performance. TabICLv2 makes the same reduction
above $700$k rows. For the six datasets that timed out at rung~$1$, we increased
only the budget at rung~$2$. At $13{,}324$\,s and $14{,}772$\,s per fold, nine sequential fits
exceeded available budgets for two single-split datasets. Rung~$3$ caps their support at $500{,}000$ rows,
down from $814{,}751$ and $1{,}000{,}000$ for the scored refit ($712{,}907$ and
$875{,}000$ per bag fold). We retain one row per class and sample the rest
uniformly with a fixed seed. Both results are marked as using reduced support.
Omitting them would trigger leaderboard imputation.

\paragraph{Bagging across jobs.} BeyondArena assigns one budget to the whole
bagged fit, and AutoGluon allocates about $\text{limit}/(\text{folds}+1.2)$ per fold.
A large fp32 fold can exceed this share ($20{,}549$\,s versus $8{,}889$\,s on
amex\_non\_iid\_1m), triggering AutoGluon's deadline check. A sufficient budget
would require about $50$\,h in one job. We therefore run folds separately and
merge the results. Out-of-fold predictions sum exactly because each fold's
array is zero outside its validation rows. Seeding folds by absolute index
preserves splits and seeds. A test reproduced the single-job result exactly
($0.30880966400279025$). Only the budget changes, as each fold can use the full
limit. We used this procedure for eight datasets, namely
amex\_non\_iid\_1m, consumer\_complaints\_1m, home\_credit\_default\_risk,
home\_credit\_default\_stability\_1m, ieee\_fraud\_detection, lending\_club\_1m,
porto\_seguro, and pva\_revenue\_prediction\_kddcup98. Of these, only
porto\_seguro uses rung~$0$. The other $87$ rung~$0$ datasets run in one job.

\section{Pretraining Priors}
\label{app:priors}

This section summarizes the distributions used in \cref{sec:experiments}
without giving full generation details.

\paragraph{Structural-causal-model prior.} The synthetic prior follows
\citet{muller2022pfn,qu2025tabicl,qu2026tabiclv2}. We propagate noise through a
sampled directed acyclic graph with assigned node mappings. Selected nodes
supply features, and one supplies the target, discretized for classification.
Table shape and labeled fraction vary across draws. We generate tables on the
fly without reuse, epochs, or a validation split.

\paragraph{Self-supervised prior over real tables.} Stages~B and~C mix real
and synthetic episodes in equal proportions, following
\citet{ma2025tabdpt,arazi2025mitra}. We use TabDPT's curated corpus of public
OpenML tables. Removing every table that overlaps with TabArena or BeyondArena reduces the corpus from $123$ to $108$ tables and keeps evaluation
data out of pretraining.

In each episode, we mask one column and use it as the target. The remaining
columns are features. We bin continuous targets into classes using random
value thresholds and retain the codes of low-cardinality targets. We shuffle
and subsample the feature columns to obtain multiple tasks from each table.
Rows are sampled uniformly without replacement and split into support and
query sets. We use the same loss as for synthetic tables.

We fit feature moments on support rows alone to avoid leaking query statistics
into the context. The reference implementation fits them once per table.
We also omit retrieval. The upstream implementation builds support from a
query row's nearest neighbours, forming a local cluster with low class
diversity. Our serving path reads the full support.

\end{document}